# Hybrid Wave-wind System Power Optimisation Using Effective Ensemble Covariance Matrix Adaptation Evolutionary Algorithm

Mehdi Neshat[1,2*], Nataliia Y. Sergiienko[2], Leandro S.P. da Silva[3], Seyedali Mirjalili[4,5], Amir H. Gandomi[1,5], Ossama Abdelkhalik[6] and John Boland[7*]

[1] *Faculty of Engineering and Information Technology, University of Technology Sydney, Ultimo, Sydney, 2007, NSW, Australia*
gandomi@uts.edu.au, mehdi.neshat@uts.edu.au

[2] *School of Electrical and Mechanical Engineering, University of Adelaide, Adelaide, SA, 5005, Australia* nataliia.sergiienko@adelaide.edu.au

[3] *Delmar Systems, Perth, WA, 6000, Australia* ldasilva@delmarsystems.com

[4] *Center for Artificial Intelligence Research and optimisation, Torrens University Australia, Brisbane, QLD 4006, Australia*
ali.mirjalili@torrens.edu.au

[5] *University Research and Innovation Center (EKIK), Obuda University, Budapest, 1034, Hungary*

[6] *Department of Aerospace Engineering, Iowa State University, Ames, IA, 50011, United States* ossama@iastate.edu

[7] *Industrial AI Research Centre, UniSA STEM, University of South Australia, Mawson Lakes, 5095, Australia*
John.Boland@unisa.edu.au

**Abstract**

Floating hybrid wind-wave systems combine offshore wind platforms with wave energy converters (WECs) to create cost-effective and reliable energy solutions. Adequately designed and tuned WECs are essential to avoid unwanted loads that could disrupt turbine motion while efficiently harvesting wave energy. These systems diversify energy sources, enhancing energy security and reducing supply risks while providing a more consistent power output by smoothing energy production variability. However, optimising such systems is complex due to the physical and hydrodynamic interactions between components, resulting in a challenging optimisation space. This study uses a 5-MW OC4-DeepCwind semi-submersible platform with three spherical WECs to explore these synergies.

To address these challenges, we propose an effective ensemble optimisation (EEA) technique that combines covariance matrix adaptation, novelty search, and discretisation techniques. To evaluate the EEA performance, we used four sea sites located along Australia's southern coast. In this framework, geometry and power take-off parameters are simultaneously optimised to maximise the average power output of the hybrid wind-wave system. Ensemble optimisation methods enhance performance, flexibility, and robustness by identifying the best algorithm or combination of algorithms for a given problem, addressing issues like premature convergence, stagnation, and poor search space exploration. The EEA was benchmarked against 14 advanced optimisation methods, demonstrating superior solution quality and convergence rates. EEA improved total power output by 111%, 95%, and 52% compared to WOA, EO, and AHA, respectively. Additionally, in comparisons with advanced methods, LSHADE, SaNSDE, and SLPSO, EEA achieved absorbed power enhancements of 498%, 638%, and 349% at the Sydney sea site, showcasing its effectiveness in optimising hybrid energy systems.

* Corresponding author

List of Abbreviations and Nomenclature

| Abbreviation | Description |
|---|---|
| WEC | Wave Energy Converter |
| MARINA | Marine Renewable Integrated Application Platform |
| MERMAID | Innovative Multi-Purpose Offshore Platforms: Planning, Design, and Operation |
| ORECCA | Offshore Renewable Energy Conversion Platforms Coordination and Support Action |
| WEC | Wave Energy Converter |
| FOWT | Floating Offshore Wind Turbine |
| PTO | Power Take-Off |
| LCOE | Levelised Cost of Energy |
| $B_{pto}$ | PTO damping coefficient |
| $K_{pto}$ | PTO stiffness coefficient |
| GA | Genetic Algorithm |
| NS | Novelty Search |
| PSO | Particle Swarm Optimisation |
| MPA | Marine Predators Algorithm |
| DE | Differential Evolution |
| CMA-ES | Evolution Strategy with Covariance Matrix Adaptation |
| GWO | Grey Wolf Optimiser |
| WOA | Whale Optimisation Algorithm |
| AHA | Artificial Hummingbird Algorithm. |
| EO | Equilibrium Optimiser |
| SCA | Sine Cosine Algorithm |
| ESO | Enhanced Snake Optimiser |
| Variable Name | Description |
| $\alpha$ | Sine-cosine control parameter (used in SCA and WOA) |
| Iter | Current iteration number |
| $Max_{Iter}$ | Maximum number of iterations |
| $\sigma_{qi}$ | Standard deviation of the relative velocity |
| $\omega$ | Angular frequency (rad/s) |
| $q_n$ | Wind turbine nacelle acceleration |
| $P_{hybrid}$ | Hybrid system power output |
| $z_{WT}$ | Vertical coordinate of the wind turbine nacelle |
| $z_{cg}$ | Vertical coordinate of the platform's center of gravity |
| $\sigma_{qWT}$ | Standard deviation of nacelle acceleration |
| $O(H_s, T_p, U_w)$ | Probability of occurrence of environmental conditions |
| $\sigma_k$ | Step size in CMA-ES |
| $C_k$ | Covariance matrix in CMA-ES |
| $N_s$ | Neighborhood size for novelty search |
| $dist(x, \lambda_i)$ | Euclidean distance between $x$ and $\lambda_i$ |
| $q$-factor | Quality factor of the WEC system |
| $H_s$ | Significant wave height |
| $T_p$ | Peak wave period |
| $U_w$ | Wind speed |
| φ | Acceleration coefficients (used in PSO) |
| χ | Constriction coefficient (used in PSO) |
| λ | Offspring count (used in CMA-ES) |
| μ | Parent count (used in CMA-ES) |

Table 1: List of abbreviations and nomenclature used in the paper.

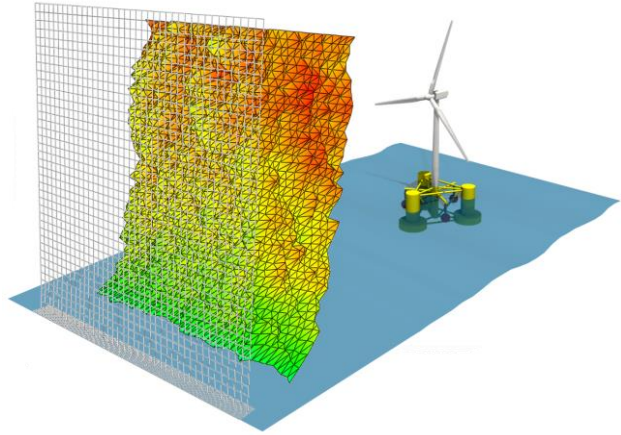

Figure 1: Hybrid offshore renewable energy platform composed a FOWT and three PAs, and environmental characteristics including wind and wave loading.

1. **Introduction**

Renewable energy sources can play an important role in meeting the growing global demand for energy with a cleaner, more sustainable, and environmentally friendly solution [1, 2]. Among the various types of renewable energy sources, wind and wave energy are receiving increasing attention worldwide due to their promising potential for large-scale electricity generation [3, 4]. Wind energy conversion systems have been widely implemented throughout the world, with proven efficiency, sustainability, and economic viability in numerous countries [5]. However, the drawback lies in its intermittency, making it a less reliable source of energy [6], as it is highly dependent on the speed and consistency due to geographical location, season, and time of day [7, 8]. On the other hand, the energy harnessed from ocean waves is a relatively untapped resource with substantial potential that can provide a constant power supply. However, wave energy converters have technical and logistical challenges associated with deployment, operation, and maintenance, hindering their widespread adoption [9, 10]. To circumvent individual technological challenges, an increasing number of papers have investigated the deployment of wave-wind energy systems to reduce dependence on a single technology or fuel source, thereby increasing energy security and reducing the risk of supply disruptions [11].

The concept behind merging wave-wind energy systems lies in its ability to compensate for the shortcomings of one renewable energy source with the other. This will result in more consistent and reliable energy output, smoothing the variability in energy production and providing a dependable power source [12]. In addition, according to a recent study conducted by Wave Energy Scotland [13], it is possible to reduce the levelised cost of energy for both wind and wave energy technologies by sharing projects and infrastructure. The combined benefit is found to be around 12%, the cost reduction is up to 40% for wave energy, while it is up to 7% for wind. The combination of both technologies can be done just by co-locating wind and wave devices or by installing wave energy converters on the same offshore platform as the wind turbine, forming a hybrid wind-wave system. There are several benefits in integrating both energy sources, such as sharing the costs, supply chain and manufacturing, and smoothing the power supply [14], and providing a stable, constant, and diversified supply of energy, while increasing the commercialization of wave energy converters. Co-located wind and wave systems have less dynamic interaction as compared to hybrid platforms, which makes the design and integration easier and less risky [15]. However, combining wave energy converters (WECs) and the floating platform in a single platform (hybrid system), such as the one shown in Figure 1, may lead to higher cost savings and improvements in the system's dynamics.

In this general context, industry and academia have been investigating and proposing hybrid platforms, including complex research projects, such as the ORECCA [16], H2OCEAN [17], TROPOS [18], MARINA [19], MERMAID [20], and Australia-China JRC of Offshore Wind and Wave Energy Harnessing [21, 22]. Among the combinations, more than 40 types [23] of floating hybrid wind-wave platforms have been proposed and analysed, including many wind turbine platforms (spar, semi-sub, tension-leg, etc.) and WEC types (oscillating buoys, water columns, surge converters, etc.). A common practice for developing a hybrid structure is relying on existing devices and tuning the WEC parameters to improve the system response [24]. More than 70% of hybrid solutions build existing floating platform designs and add wave power capabilities to them, while approximately 30% of publications are dedicated to completely new platforms for installing one or more wind turbines and wave power devices. This approach is usually investigated due to the reduced number of variables to design and optimise the structure, offering an idea of the benefits of combining both devices.

In this general context, several authors performed sensitivity and optimisation studies on hybrid platforms to understand the behaviour and additional benefits of such platforms. When considering the WEC side, the main design variables of such a system are the shape, size, distance to the FOWT, and number of WECs [25–28]. In addition, the hybrid platform is highly dependent on the power take-off configuration and control algorithms chosen to tune the WEC motion. Control strategies that are designed to maximise WEC power might cause undesirable motions of the floating foundation [29, 30] and excessive nacelle acceleration, which in turn increases the load on the tower and reduces its fatigue life [31, 32]. Therefore, the use of WECs to suppress the motion [33–35] and to potentially reduce the tower root stress [36, 37] can provide additional potential benefits for the hybrid wind-wave system. It is important to note that even though the nacelle acceleration can be reduced, some concepts lead to an increased heave motion, which impacts the mooring loads [24, 36]. Recent investigations under numerical and experimental analysis demonstrate that depending on the damping magnitude, the heave oscillation can be intensified [28], while reasonable damping can suppress the pitch motion. In addition, it was demonstrated that the WEC control for power increment, motion reduction, and load mitigation (tower base) can be simultaneously obtained [28]. Therefore, future work should investigate mitigating undesired motions while optimising ratios of the WECs and FOWTs [14]. Research gaps and future perspectives highlight the need for an optimisation/framework of the hybrid platform from a global geometric point of view, where the conceptual design should be investigated, focusing on the dynamic behaviour (motion response and power absorption), followed by the structural response and control strategies under operational conditions [23]. However, designing a hybrid platform can be a hard and intensive task due to the large number of possible hybrid platform configurations and objective function options. The challenge relies on the complex interplay of physical and hydrodynamic interactions between devices, which makes optimisations more challenging when compared to individual wind or wave energy systems. These interactions result in a complex optimisation space with heterogeneous parameters, complicating the search for optimal solutions [38, 39].

Maximising power output from hybrid platforms requires fine-tuning diverse parameters from wind turbines and WECs, making the problem highly multi-modal and computationally demanding [40]. Among a wide range of optimisation techniques, nature-based methods like evolutionary [41] and swarm intelligence algorithms [42] show considerable performance and have been employed to address these challenges. Despite progress, a critical need remains for more robust algorithms capable of effectively handling heterogeneous characteristics and delivering high-quality solutions to meet industry demands.

One of the preliminary works [43] evaluated the impact of array layout on the power of WECs, utilising a combination of the standard genetic algorithm (GA) and parabolic crossover methods to optimise the layout. Subsequently, literature [44] applied GA to optimise WEC array layouts based on a hydrodynamic model while accounting for array costs. Additionally, reference [45] employed the differential evolution algorithm for multi-parameter optimisation of a square WEC array, demonstrating notable stability in the optimisation process.

As optimising hybrid systems is computationally expensive, a GPU-based solution [46] (multi-objective differential evolution algorithm) was proposed to optimise the layout of these systems. Compared to wind-only farms, the optimised hybrid farm achieves a 38% increase in energy output and a 44% reduction in wave loads on wind turbine foundations, and the GPU-based codes run more than 1000 times faster than CPU-based ones. To tackle the complexity of optimising hybrid wave-wind energy systems, an adaptive bi-level whale optimisation algorithm (AWOA) [47] was developed. The upper level optimises geometry parameters, while the lower level focuses on PTO parameters. Results demonstrate that AWOA effectively addresses the challenges of bi-level optimisation, confirming its efficiency.

A recent study [48] introduced an enhanced snake optimiser (ESO) featuring chaotic initialisation, asynchronous learning factors, and Levy flight. It was designed to improve optimum searching capabilities while avoiding local optima. ESO demonstrated good performance in optimising converters, numbering more than six; however, it was not compared with advanced optimisation methods. In another application of meta-heuristics in hybrid model enhancements, Yang et al. [49] introduced a hybrid algorithm (EMCO) recently. EMCO integrates an artificial ecosystem (AEO) and manta ray foraging algorithm (MRFO) to balance local and global explorations. Experiments with 3, 5, 8, and 20 hybrid models validated EMCO's effectiveness against five benchmark algorithms, and it achieved the q factor values across scales above one. Furthermore, Li et al. [50] enhanced the original slime mould algorithm (SMA) into the exponential one (ESMA) by integrating a chaos algorithm, exponential asynchronous factor, and sine-cosine mechanism to optimise power output. ESMA, designed for optimal power configuration, was tested on hybrid systems with various buoy numbers and compared with six other algorithms. Results confirm ESMA's superior performance, with its advantages becoming more evident as system size increases.

The main question arising for such a complex system is how to combine both structures and what are the most important parameters to be selected as design variables and objective functions, and their effects on the system response. Based on the above, the following paper aims to address the optimisation of a hybrid wind-wave system and demonstrate the effects of WEC size and distance from the FOWT while optimising the PTO parameters to achieve mutual benefits for a specific offshore location. The major contributions of this paper are as follows.

- An efficient ensemble optimisation method is developed that leverages the strengths of three meta-heuristic algorithms. These algorithms are dynamically combined and executed using an adaptive strategy based on their performance, significantly enhancing the total power output of a hybrid wave-wind energy system.

- To address premature convergence and local optima, common challenges in optimization methods, we propose incorporating Novelty Search to enhance exploration and maintain diversity in the search space, uncovering new and potentially valuable solutions. In real-world industrial scenarios, the best-found solution may often be impractical due to natural constraints. Novelty Search ensures a diverse population of high-quality, practical solutions, providing flexibility and robustness for implementation.

- To accelerate the exploration phase, we implement a simple discretisation strategy that initiates the optimisation process within a hyperspace. By transforming the search space into a logarithmic scale (e.g., $\log_{10}(x)$), the optimiser operates in a compressed and discretised domain, enhancing its ability to explore efficiently and converge more effectively.

- Additionally, the research conducts an in-depth comparative assessment to ascertain the efficacy of the suggested ensemble optimisation framework aimed at improving hybrid wave-wind energy systems in comparison to 14 cutting-edge optimisation techniques. This meticulous analysis underscores the merits and limitations of different methodologies, showcasing the exceptional advantages of the proposed framework.

- Lastly, the intricate and detailed optimisation of hybrid wave-wind energy systems is being undertaken for four deployment locations along Australia's South Coast. These particular sites, each exhibiting a distinct set of wave characteristics that contribute to their unique environmental conditions, diligently address the specific optimisation requirements that are essential for each location. In doing so, the research ensures that the proposed framework is not only robust but also remarkably adaptable to the various and often unpredictable sea conditions that may arise in this diverse maritime landscape.

The paper is organised as follows. We present a comprehensive system description and modelling of the hybrid wave-wind system (section 2), detailing its physical and hydrodynamic interactions. The methodology presented in Section 3, elucidates the meta-heuristic optimisation algorithms that are employed, with a particular emphasis on the Novelty Search (NS) and the ensemble optimisation algorithm. The efficacy of our proposed method is then evaluated and discussed in section 4, where it is compared against various renowned optimisation techniques. Ultimately, we draw our conclusions in section 5, where we summarise the findings of our study,

## 2. System description and modelling

### 2.1. Hybrid wind-wave system

The considered hybrid wind-wave system consists of the OC4-DeepCwind semi-submersible platform supporting a 5-MW wind turbine [51] and three floating half-submerged spherical WECs as shown in Figure 2. The choice of technologies for hybridisation has been driven by the fact that these particular FOWT and WECs have been extensively studied (separately) numerically and experimentally [52], and their dynamics and power production (for WECs) are well-understood.

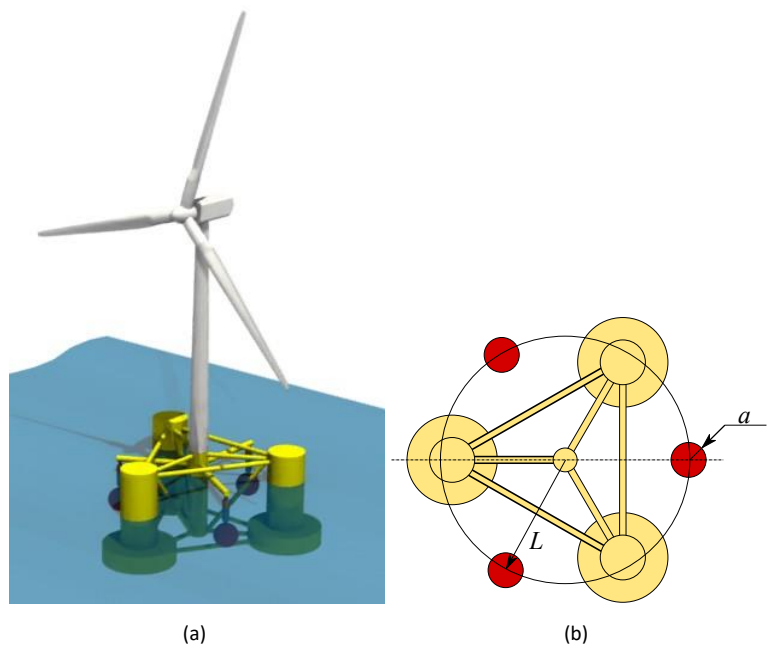

(a)　　　　　　　　　　　(b)

Figure 2: Hybrid wind-wave energy system consisting of a 5-MW DeepCwind-OC4 floating wind platform combined with three half-submerged spherical WECs: (a) artist's impression of the system, (b) op view of the hybrid wind-wave system (the floating platform is shown in yellow, and three spherical WECs are shown in red). $a$ is the WECs' radius, and $L$ is the distance between the center lines of the platform and the WECs.

Two well-studied technologies are chosen for investigation and design optimisation in this study. The wind power generation system is a semi-submersible platform [51] designed in Phase II of the Offshore Code Comparison Collaboration Continuation (OC4) project to support a 5-MW offshore baseline wind turbine [53]. The wave energy system consists of three half-submerged spherical WECs that are located around the platform's central column. The WEC radius is set to 5 m as this geometry has been widely used as a benchmark for modelling verification and validation [54], control system development [55], etc. It is assumed that the WECs are connected to the platform via three power take-off (PTO) units allowing WECs to convert wave power into electricity and to adjust the motion of the platform (if needed). The distance from the platform's centre to the WECs is kept as an optimisation variable as well as the WEC PTO parameters.

The radius of WECs and their distance from the central column are used as optimisation variables. The mass of each WEC is set to the mass of the displaced water (so the WECs are neutrally buoyant), which was assumed to reduce the effects of a buoy loss on platform stability. It is assumed that each WEC is connected to the power take-off (PTO) machinery, the dynamics of which can be approximated by the spring-damper system. Dimensions, inertia properties, and mooring line characteristics of the semi-submersible platform are taken from [51] and remain unchanged.

## 2.2. Dynamic model

The hybrid wind-wave system is subjected to hydrodynamic loading on the floating platform and WECs, and aerodynamic loading on the wind turbine, described in Appendix A and based on an in-house model using MATLAB. The modelling methodology is outlined in Figure 3. The analysis is conducted in the frequency domain, where important sources of nonlinearities are included using statistical linearisation; for more information, the reader is encouraged to read the material presented in [24, 56]. The hydrodynamic coefficients consider multi-body interactions and were calculated using WAMIT [57]. Note that second-order forces from waves were not considered in this study due to the computational cost when investigating a multi-body system and several geometries during an optimisation procedure.

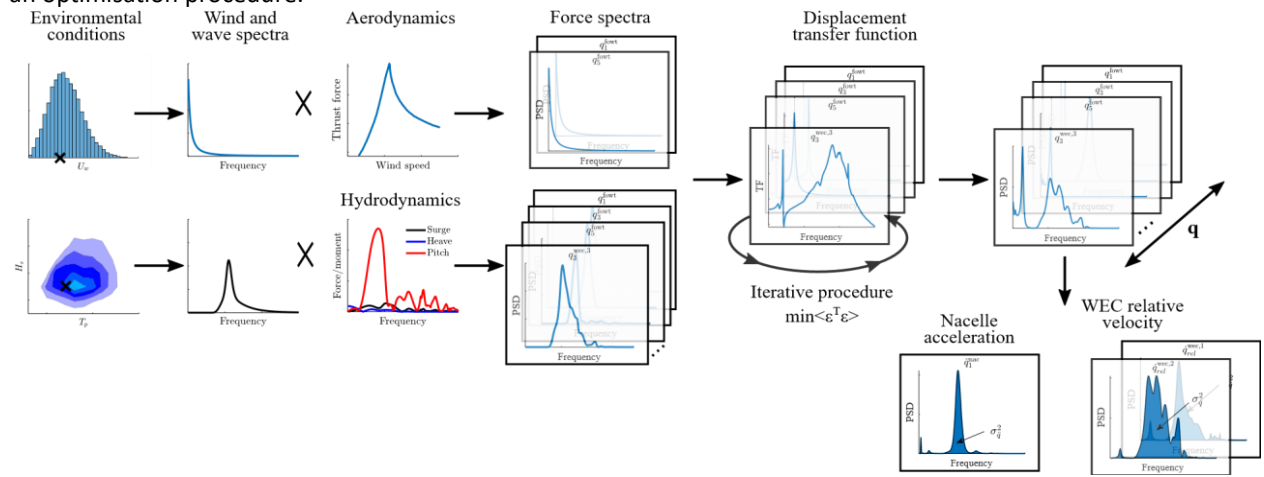

Figure 3: Outline of the methodology used for modelling a hybrid wind-wave energy system in spectral domain.

In addition, second-order effects are assumed to be minor in this analysis as the response investigated is given in the wave frequency range, such as power absorbed, heave motion, and nacelle acceleration, which are considered to be more dependent on the wave frequency loads, where the WEC coupling is more relevant.

Irregular waves and turbulent winds represent environmental loading, and both sources are assumed to be colinear and coming from 0deg (facing the platform). Note that in real sites, wave-spreading and non-colinear environmental loads affect the response of the platform and its natural frequency [58]. The dynamics of the hybrid wind-wave system are described by eight states: the platform's motion is represented by surge $q_1^p$, heave $q_3^p$ and pitch $q_5^p$ amplitudes, and the WECs' dynamics is represented by their heaving motion $q^w{}_3{}^{,i}, i = 1...3$.

*2.3. Equations of motion*

The hybrid wind-wave system dynamics is modelled in the spectral domain using eight generalised coordinates: five for the floating wind system and three for the wave energy system. Assuming that the incident waves and wind speeds are colinear in this study, the dynamics of the floating offshore wind turbine (FOWT) contain three rigid-body motions of the semi-submersible platform ($p$), namely surge $q_1^p$, heave $q_3^p$ and pitch $q_5^p$, and two states that describe the wind turbine, the collective blade pitch angle $\beta$ and the generator speed $\Omega = \dot{\varphi}$. The wave energy system is described by three heaving motions of the WECs as $q_3^{w,i}, i = 1...3$. As a result, the state vector of the hybrid wind-wave system is taken as:

$$\mathbf{q}(\omega) = \left[q_1^p, q_3^p, q_5^p, \phi, \beta, q_3^{w,1}, q_3^{w,2}, q_3^{w,3}\right]^\mathrm{T} \tag{1}$$

The resultant equation of motion of the hybrid wind-wave system is written as:

$$\mathbf{M}\ddot{\mathbf{q}}(t) = \mathbf{F}_{\mathrm{exc}}(t) + \mathbf{F}_{\mathrm{rad}}(t) + \mathbf{F}_{\mathrm{aero}}(t) + \mathbf{F}_{\mathrm{moor}}(t) + \mathbf{F}_{hs}(t) + \mathbf{F}_{visc}(t) + \mathbf{F}_{\mathrm{pto}}(t), \tag{2}$$

where the dominant forces acting on the hybrid system include (i) hydrodynamic loading on the floating platform and three WECs that consist of the wave excitation $F_{exc}(t)$, radiation $F_{rad}(t)$ and hydrostatic forces $F_{hs}(t)$; (ii) aerodynamic loading $F_{aero}(t)$ on the wind turbine is represented by the rotor thrust force and generator torque; (iii) quadratic viscous damping loading $F_{visc}(t)$ on the platform and the WECs; (iv) mooring forces on the floating platform $F_{moor}$; and (v) WEC PTO forces $F_{pto}(t)$ that couple the dynamics of the floating platform and three WECs.

The nonlinear effects included in Equation (2), such as aerodynamic forces, viscous drag, and hydrostatic forces, are linearised following the statistical linearisation approach [56] and represented by the equivalent damping and stiffness terms. Therefore, the linearised equation of the hybrid wind-wave system represented in the frequency domain has a form:

$$\left[-\omega^2(\mathbf{M} + \mathbf{A}_{\mathrm{rad}}) + i\omega\left(\mathbf{B}_{\mathrm{rad}} + \mathbf{B}_{\mathrm{aero}}^{eq} + \mathbf{B}_{\mathrm{visc}}^{eq} + \mathbf{B}_{\mathrm{pto}}\right) + \mathbf{K}_{\mathrm{moor}} + \mathbf{K}_{hs}^{eq} + \mathbf{K}_{pto}\right]\mathbf{q}(\omega) = \mathbf{Q}(\omega), \tag{3}$$

where M is the mass matrix of the multi-body system (platform mass properties are taken from [51] and the WEC's mass is evaluated based on the submerged geometry), hydrodynamic added mass $A_{rad}$ and radiation damping matrices $B_{rad}$ are evaluated using WAMIT [59], mooring stiffness matrix $K_{moor}$ is evaluated following the quasi-static approach detailed in [60, 61], $B_{pto}$ and $K_{pto}$ are the PTO coupling matrices, and Q is the generalised load vector that contains the wave excitation force and wind loads. All matrices with a superscript 'eq' correspond to the equivalent linear matrices that are calculated iteratively for each wind and wave condition following statistical linearisation (for more details refer to [62]).

*2.4. Deployment sites*

Recently, a number of studies have explored the economic benefits of combining wind and wave energy systems in Australia [63–66]. The information about four offshore sites in different states is presented in Table 2, and their statistical data is shown in Figure 4.

Table 2: Information about four offshore sites around Australia [65], statistical data for the period between 2014 and 2020.

| Location | State | Coordinates | Depth | Distance | Wave power | Wind power |
|---|---|---|---|---|---|---|
| Sydney | NSW | 34.0°S 152.3°E | >100 m | 117 km | 19.7 kW/m | 1.07 kW/m² |
| Gippsland | VIC | 38.9°S 146.9°E | 25 m | 16 km | 4.5 kW/m | 0.67 kW/m² |
| Port Lincoln | SA | 34.9°S 135.5°E | 70 m | 10 km | 51.7 kW/m | 0.68 kW/m² |

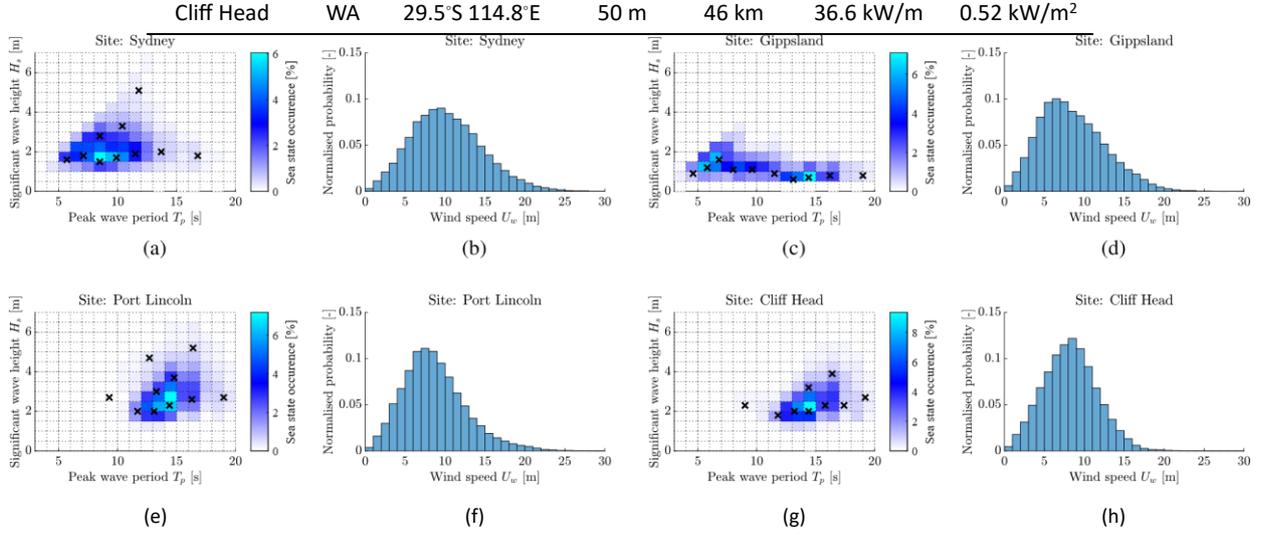

Figure 4: Wind and wave climate statistics for offshore sites in Australia from 2014 to 2020.

## 2.5. Performance measures

The performance of the hybrid wind-wave system is assessed against two key parameters: the power production of three WECs, and wind turbine nacelle acceleration. The average power generated by three WECs in one sea state is evaluated as [62]:

$$\bar{P}^w = \sum_i B_{pto,i} \left(\sigma(\dot{q}_{3rel}^{w,i})\right)^2, \tag{4}$$

where $B_{pto,i}$ is the PTO damping coefficient associated with $i^{th}$ WEC, and $\sigma(\dot{q}^w{}_3{}^{,i})$ denotes the standard deviation of the relative velocity between the $i^{th}$ WEC and the platform. The relative motion is given by:

$$q_{3rel}^{w,i} \approx q_3^{w,i} - q_3^p + q_5^p x^{w,i} \tag{5}$$

where $x^{w,i}$ is the $x$ coordinate of the WEC's centre of gravity, which a variable investigated in this work.

The wind turbine performance is characterised by the horizontal component of the wind turbine horizontal nacelle acceleration [35]:

$$\ddot{q}_1^n = \ddot{q}_1^p + \ddot{q}_5^p (z_n - z_{cg}), \tag{6}$$

where $z_n$ and $z_{cg}$ are vertical coordinates of the nacelle and the platform's centre of gravity. The nacelle acceleration fluctuations are represented using its standard deviation for each environmental condition of interest. Note that the tower flexibility was not included in this work, as this is part of an initial assessment, and different tower configurations can be designed, from the Soft-Soft up to Stiff-Stiff region of the first tower bending mode [67].

Once each environmental condition from Eq 7 using $H_s$, $T_p$, and $U_w$ are calculated, the annual average power production and average standard deviation of the nacelle acceleration are evaluated for the Sydney site using the probability of occurrence $O(H_s,T_p,U_w)$ from Eq 7:

$$\tilde{P}^w = \sum_{10} O(H_s, T_p, U_w) \cdot \bar{P}^w(H_s, T_p, U_w), \tag{7}$$

$$\tilde{\sigma}(\ddot{q}_1^n) = \sum_{10} O(H_s, T_p, U_w) \cdot \sigma(\ddot{q}_1^n)(H_s, T_p, U_w). \tag{8}$$

In a similar way, the average response of the platform in surge $\tilde{\sigma}(\dot{q}_1^p)$, heave $\tilde{\sigma}(\dot{q}_3^p)$ and pitch modes $\tilde{\sigma}(\dot{q}_5^p)$ are evaluated at the selected deployment site.

The hybridisation of wave and wind systems should be mutually beneficial for both technologies, therefore, the two main performance measures are considered: power production of the wave energy system, and nacelle acceleration of the wind turbine that is used as a proxy for wind turbine loads.

The average power generation of the wave energy system in one sea state is evaluated as [62]:

$$P_{hybrid}^{WEC}(H_s, T_p, U_w) = \sum_i \sigma_{\dot{q}^i}^2 B_{pto,i}, \tag{9}$$

where $B_{pto,i}$ is the PTO damping coefficient associated with the $i^{th}$ WEC, and $\sigma_{\dot{q}^i}$ denotes the standard deviation of the relative velocity between the $i^{th}$ WEC and the platform. Once Equation (A.20) is solved with respect to q(ω), it is possible to evaluate the standard deviation of its response in each degree of freedom.

The average annual power production of WECs in a hybrid system is evaluated as:

$$\bar{P}_{hybrid}^{WEC} = \sum_{H_s, T_p, U_w} P_{hybrid}^{WEC}(H_s, T_p, U_w) \cdot O(H_s, T_p, U_w), \tag{10}$$

where $O(H_s, T_p, U_w)$ is the probability of occurrence of each environmental condition.

The wind turbine performance is characterised by the horizontal component of the wind turbine nacelle acceleration [31, 56]:

$$\ddot{q}_{hybrid}^{WT} = \ddot{q}_1^p + \ddot{q}_5^p(z_{WT} - z_{cg}), \tag{11}$$

where $z_{WT}$ and $z_{cg}$ are vertical coordinates of the nacelle and the platform's centre of gravity. As nacelle acceleration fluctuates around zero, it is more representative to calculate its standard deviation $\sigma_{\ddot{q}^{WT}_{hybrid}}$ for each environmental condition of interest. Then, the annual average nacelle acceleration of the wind turbine at the given deployment site is evaluated as:

$$\bar{\sigma}_{\ddot{q}_{hybrid}^{WT}} = \sum_{H_s, T_p, U_w} \sigma_{\ddot{q}_{hybrid}^{WT}}(H_s, T_p, U_w) \cdot O(H_s, T_p, U_w). \tag{12}$$

The ratio $\sigma_{\ddot{q}\text{-}WThybrid}/\sigma_{\ddot{q}\text{-}WTisolated}$ can provide an understanding whether the WECs are used to decrease the loading on the wind turbine in the hybrid wind-wave system. Here $\sigma_{\ddot{q}\text{-}WTisolated}$ is evaluated as the annual average nacelle acceleration of the wind turbine that is installed on a floating platform in the absence of WECs.

## 3. Optimisation methodology

Evolutionary and swarm optimisation methods have been extensively applied to ocean wave energy systems to enhance energy collection efficiency and reliability [68]. These optimization methods, such as Genetic Algorithms (GA), Particle Swarm Optimization (PSO), and Differential Evolution (DE), are most suitable for control strategy design and optimisation of wave energy converters (WECs) in highly dynamic and nonlinear marine environments [69]. Parameter modification such as device geometry, power take-off systems [70], and placement configurations maximize energy collection with regard to environmental variability and operational constraints.

### 3.1. Novelty Search (NS)

Novelty search [71] is a concept in evolutionary computation that emphasises the significance of exploring and rewarding novel solutions while searching for an optimal solution. In traditional optimisation algorithms, the fitness function assesses candidate solutions, and the search concentrates on discovering solutions with high fitness values. Nevertheless, in some cases, the search may evolve trapped in local optima or need help exploring whole feasible solutions. On the contrary, novelty search [72] focuses on exploring the diversity of the search space to uncover renewed and potentially valuable solutions that may hold low fitness values. In this regard, a novelty metric should be defined to evaluate the candidate solution novelty degree depending on how diverse it is from earlier experienced explanations. Next, the novelty metric guides the search, intending to discover the most diverse solutions that can lead to new insights or breakthroughs in the problem domain. Likewise, by sustaining diverse solutions, the optimisation algorithm may avoid getting trapped in local optima and continue exploring the full range of the search space. Novelty search has been applied in various disciplines, including robotics [73], game design [74], and machine learning [75]. It has been shown to be effective in cases where traditional fitness-based optimisation methods fail to produce optimal solutions or get stuck in local optima [76]. However, novelty search can also be computationally expensive and may demand further parameter tuning compared to traditional optimisation methods. The novelty metric is computed as explained in Equation. 13 in this study.

$$p(x) = \frac{1}{N_s} \sum_{i=1}^{N_s} \text{dist}(x, \lambda_i) \tag{13}$$

where $\lambda_i$ is the $i^{th}$ most adjacent neighbour of $x$ in neighbourhood search space (or global), and the function of *dist* is a Euclidean distance formula. $N_s$ sets as six, which is the neighbourhood rate.

### 3.2. Artificial Hummingbird Algorithm

The AHA (Artificial Hummingbird Algorithm) algorithm [77] is designed to mimic the unique flight skills and intelligent foraging strategies observed in hummingbirds. It incorporates three types of flight skills: axial, diagonal, and omnidirectional flights, which are integral to the hummingbirds' foraging strategies. The algorithm also implements guided foraging, territorial foraging, and migrating foraging while incorporating a visiting table to emulate the memory function of hummingbirds in remembering food sources. The effectiveness of AHA is validated through numerical test functions, surpassing various other meta-heuristic algorithms in terms of competitiveness and generating high-quality solutions with fewer control parameters. Furthermore, AHA's performance is assessed through ten challenging engineering design case studies, where it demonstrates superior effectiveness in terms of computational burden and solution precision compared to existing optimization techniques documented in the literature.

### *3.2.1. Guided foraging*

In the AHA (Artificial Hummingbird Algorithm), the foraging behaviour of particles is guided by a natural tendency to visit food sources with maximum benefits. This is achieved by assigning a high interest rate and prolonging the unvisited time for target sources. Once the target food source is determined, a particle can navigate towards it for feeding. The AHA algorithm incorporates three flight skills: omnidirectional, diagonal, and axial flights, which are modelled using a direction switch vector. The direction switch vector controls the availability of directions in the dimensional space. Axial flight allows particles to move along any coordinate axis; diagonal flight enables movement between opposite corners of a rectangle using any two coordinate axes out of three, while omnidirectional flight allows projection to each of the three coordinate axes. While all birds use an omnidirectional flight, only hummingbirds possess the ability to perform axial and diagonal flights. These flight patterns are extendable to higher-dimensional spaces, where axial flight is defined as follows:

$$D^{(i)} = \begin{cases} 1 & \text{if } i = \text{randi}([1, d]) \\ 0 & \text{else} \end{cases} i = 1, \ldots, d \qquad (14)$$

Equation 15 shows the diagonal movement as follows:

$$D^{(i)} = \begin{cases} 1 & \text{if } i = P(j), j \in [1, k], P = \text{randperm}(k), k \in [2, \lceil r_1 \cdot (d-2) \rceil + 1] \\ 0 & \text{else} \end{cases} \qquad (15)$$

The flight abilities of a hummingbird, including axial, diagonal, and omnidirectional flights, enable it to reach its target food source and acquire a candidate food source. In the context of the AHA algorithm, the selection of a target food source triggers the updating of a food source. This update process involves considering the selected target food source in relation to all the existing sources. To mathematically model the guided foraging behaviour and obtain a candidate food source, a specific equation is derived. This equation (16) captures the dynamics and interactions involved in the foraging process, allowing for the simulation of the hummingbird's behaviour within the algorithm.

$$v_i(t+1) = x_{i,\text{tar}}(t) + \alpha \cdot D \cdot \left(x_i(t) - x_{i,\text{tar}}(t)\right) \qquad (16)$$

The position update Equation (16) in the AHA algorithm incorporates the guided foraging behaviour of hummingbirds. It involves updating the position of each food source based on its proximity to the target food source, denoted as $x_i(t)$ and $x_{i,tar}(t)$, respectively. The update is influenced by a guided factor, α, which follows a normal distribution ($N(0,1)$). This equation enables the exploration of different flight patterns and facilitates the adaptation of the algorithm to optimize the search process. The position update of the *ith* food source is formulated and can be seen in Equation 17.

$$x_i(t+1) = \begin{cases} x_i(t) & f(x_i(t)) \leq f(v_i(t+1)) \\ v_i(t+1) & f(x_i(t)) > f(v_i(t+1)) \end{cases} \qquad (17)$$

### **3.2.2. Territorial foraging**

After consuming the nectar from its target food source, a hummingbird exhibits a preference for exploring new food sources instead of revisiting existing ones. This behaviour prompts the hummingbird to move within its territory, seeking out potential candidate solutions that may outperform the current food source. To simulate this territorial foraging strategy, the AHA algorithm employs a mathematical equation (18) for modelling the local search behaviour of hummingbirds and the identification of candidate food sources. The derivation of this equation facilitates the emulation of the hummingbird's behaviour and its search for improved solutions.

$$v_i(t+1) = x_i(t) + \beta \cdot D \cdot x_i(t) \tag{18}$$

where β is a territorial factor, formulated using a normal distribution *N*(0,1).

### *3.2.3. Migration foraging*

In the AHA algorithm, a migration coefficient (γ) is introduced to address situations where a hummingbird frequently visits a region that lacks sufficient food resources ($x_{wor}$). When the number of iterations surpasses the predetermined migration coefficient value, a hummingbird positioned at the food source with the poorest nectar-refilling rate will undergo a random migration to a new food source generated across the entire search space. During this migration process, the hummingbird abandons its previous location and settles at the new food source for feeding. Subsequently, the visit table, which maintains the memory of food sources, is updated accordingly. The migration foraging behaviour of a hummingbird, transitioning from a food source with a low nectar-refilling rate to a randomly produced new source, can be described using the following equation.

$$x_{wor}(t+1) = \text{Low} + \gamma \cdot (Up - \text{Low}) \tag{19}$$

### *3.3. Ensemble optimisation algorithms*

Ensemble optimisation algorithms [78] synergistically combine multiple individual optimisation algorithms or variations of a single algorithm. By integrating a diverse set of algorithms, they can harness each component's distinct strengths and characteristics. This combination of algorithms enables them to explore a wider span of the search space, leading to enhanced coverage and heightened prospects of discovering superior solutions [79]. Furthermore, individual optimisation algorithms often manifest varying performances across different problem instances or at different stages of the optimisation process. By combining multiple algorithms, ensemble optimisation methods attain heightened robustness. In instances where one algorithm encounters difficulties or exhibits subpar performance, other algorithms within the ensemble compensate for and contribute to an overall superior performance [80]. Another notable attribute of ensemble models is their adeptness at maintaining a well-balanced exploration-exploitation equilibrium.

Typically, optimisation algorithms face a delicate trade-off between exploration, which involves seeking novel and potentially superior solutions, and exploitation, which entails refining and optimising promising solutions [81]. Ensemble optimisation algorithms excel at striking an optimal balance between exploration and exploitation by incorporating diverse strategies from their constituent algorithms. This harmonious approach impedes premature convergence to suboptimal solutions and fosters comprehensive exploration of search space.

Lastly, ensemble optimisation algorithms frequently incorporate adaptive mechanisms that dynamically adjust the contributions or weights assigned to their constituent algorithms during optimisation [82]. These adaptive mechanisms empower the ensemble to adapt to the distinct characteristics of the problem at hand. They enable the allocation of more resources to better-performing members or strategies within the ensemble and facilitate the adaptive refinement of the solution-generation process.

### *3.4. Ensemble evolutionary algorithm with novelty search*

Attaining an exceptional equilibrium between exploration and exploitation poses significant challenges within the intricate search space characterised by multiple optimal solutions arising from intricate interactions among optimisation parameters. To mitigate these challenges, we propose a novel approach called the Ensemble evolutionary algorithm (EEA), combining two distinct optimisation methods: the Whale Optimisation Algorithm (WOA) and the Artificial Hummingbird Algorithm (AHA). This synergistic fusion capitalises on the distinctive traits of each algorithm, resulting in enhanced performance and efficiency. The selection of these optimisers is primarily motivated by the unique strengths they bring to the table.

WOA demonstrates remarkable prowess in global exploration, showcasing a high convergence rate that often surpasses other optimisers during the initial stages. Leveraging random movements and position adjustments based on the best solutions discovered thus far, WOA navigates the search space with remarkable efficiency. Conversely, AHA incorporates adaptive mechanisms that enable dynamic parameter adjustments, including memory consideration, pitch adjustment rate, and improvisation. This adaptive capability empowers AHA to strike an optimal balance between exploration and exploitation. AHA is particularly well-suited for tackling complex and multi-modal search spaces, as it adapts its search strategy to effectively exploit diverse regions.

By combining WOA and AHA, our proposed approach achieves a symbiotic synergy that harnesses the strengths of both algorithms. This integration enables improved performance and efficiency in tackling the challenges of the intricate search space, allowing for practical exploration and exploitation of the optimisation landscape.

### 3.4.1. Covariance Matrix Adaptation

In this study, we combined the proposed optimisation method with Covariance Matrix Adaptation (CMA) [83] due to its remarkable proficiency in tackling intricate, high-dimensional, and inherently noisy problems while requiring only minimal adjustments. CMA is a sophisticated and exceptionally resilient optimization strategy. Its innovative approach to dynamically adapting the covariance matrix enables it to navigate the vast and complex search space with remarkable efficiency by effectively capturing the intricate correlations between variables and making necessary adjustments in response to the varying topological features of the landscape it is exploring. In a masterful display of balance, CMA-ES expertly integrates global exploration and local exploitation, thereby skillfully sidestepping the pitfalls associated with local optima and successfully attaining solutions that are truly global in nature. Its remarkable scalability and ability to leverage parallelism ensure that it operates efficiently and effectively even in large-scale and distributed environments. Particularly effective when confronted with multi-modal scenarios, CMA-ES adeptly addresses the challenges posed by noise and uncertainty, establishing itself as a versatile and dependable tool that can be applied across a broad spectrum of optimization challenges encountered in various fields and disciplines. The initial population of the solution is generated using a multivariate normal distribution (N (0,$C_k$)), where $\sigma_k$ and $C_k$ are step size and symmetric covariance matrix.

$$P_i \sim m_k + \sigma_k \times \mathcal{N}(0, C_k), i = 1, \ldots, N_\lambda \tag{20}$$

$$m_{k+1} = \sum_{i=1}^{\mu} \beta_i P_{i:N_\lambda} = m_k + \sum_{i=1}^{\mu} \beta_i (P_{i:N_\lambda} - m_k) \tag{21}$$

$$\sigma_{k+1} = \sigma_k \times \exp\left(\frac{c_\sigma}{d_\sigma}\left(\frac{\|p_\sigma\|}{\mathrm{E}\|\mathcal{N}(0,I)\|} - 1\right)\right) \tag{22}$$

$$p_\sigma \leftarrow (1 - c_\sigma)p_\sigma + \sqrt{1 - (1 - c_\sigma)^2}\sqrt{\mu_w} C_k^{-1/2} \frac{m_{k+1} - m_k}{\sigma_k} \tag{23}$$

### 3.4.2. Discretisation strategy

In light of the vast range of Power Take-Off (PTO) parameters spanning from $10^1$ to $10^{10}$, achieving effective exploration across the entire problem domain presents significant challenges. To address this issue, we propose using a discretisation technique that facilitates the transfer of continuous PTO parameters into a discretised space using the logarithm function with a base of 10. This technique effectively reduces the dimensionality of the search space and simplifies the optimisation problem by constraining the number of feasible solutions and narrowing the focus of the search to specific regions of interest. Figure 5 shows how the decision variables (PTO parameters) are discretised using the logarithm transfer function.

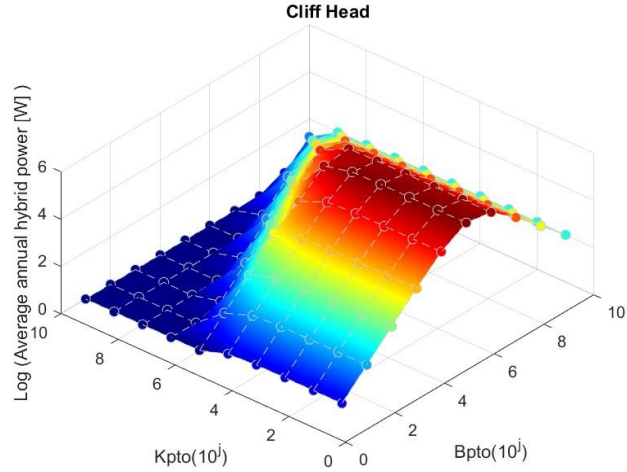

Figure 5: Descritisation landscape of PTO parameters search space for Cliff head sea site. The WECs' radius and distance are 8 m and 45 m.

By employing the discretisation technique, the optimisation algorithm gains the ability to explore and evaluate candidate solutions more efficiently within the reduced search space. The reduced dimensionality enables faster convergence towards optimal solutions by reducing the search overhead. Additionally, the computational efficiency of the optimisation process is enhanced as the discretised space allows for more straightforward calculations and faster evaluation of potential solutions.

3.4.3. Diversity consideration

In addition, we employed a novelty search strategy in conjunction with the EEA to maintain an appropriate level of solution diversity. The novelty search paradigm involves evaluating solutions based on their uniqueness or novelty rather than their fitness or performance. To measure the novelty of a solution, we implemented a comparison process using the Euclidean distance metric. By comparing each solution to others within the population or a predefined archive of previously encountered solutions, we determined their level of novelty.

To ensure the effectiveness of the novelty search strategy in enhancing solution quality, we conducted a thorough performance evaluation. We assessed its impact on the optimisation process and its ability to promote the exploration of diverse and unconventional solutions. We leveraged the novelty search approach to guide the evolutionary exploration if the evaluation indicated positive results. However, if the assessment revealed any shortcomings or limitations, we disregarded the influence of the novelty search and focused solely on other aspects of the optimisation process.

The major motivation of novelty search applications is that we aimed to strike a balance between diversity and performance in the optimisation of PTO and design parameters of the hybrid wave-wind model. This approach allowed us to explore various solutions, stimulating innovation and avoiding local optima. Through the utilisation of the Euclidean distance metric and rigorous performance evaluation, we ensured that the novelty search component positively contributed to the overall optimisation process, thereby enhancing the quality of the generated solutions. The pseudo-code of EEA can be seen in Algorithm 1.

**Algorithm 1** Ensemble evolutionary algorithm with novelty search (EEA)

1:  Initialise a randomly generated population of feasible solutions
2:  Initialise control parameters
3:  **repeat**
4:      **for** each $run$ in $R_{max}$ **do**
5:          **for** each $problem$ in $P_{max}$ **do**
6:              Generate randomly $N$ sub-particles
7:              **repeat**  ▷ Run first component (WOA)
8:                  **for** each $Particle$ in $N$ **do**
9:                      Update particle control parameters  ▷ Update $\alpha$, A, C, l, and p
10:                     **if** $p < 0.5$ **then**
11:                         **if** $|A| < 1$ **then** $\vec{D} = \left|\vec{C} \cdot \vec{X^*}(t) - \vec{X}(t)\right|$, $\vec{X}(t+1) = \vec{X^*}(t) - \vec{A} \cdot \vec{D}$
12:                         **end if**
13:                         **if** $|A| \geq 1$ **then** Select $X_{rand}$, $\vec{X}(t+1) = \overrightarrow{X_{rand}} - \vec{A} \cdot \vec{D}$  ▷ Update solutions' configuration
14:                         **end if**
15:                         **if** $p \geq 0.5$ **then** $\vec{X}(t+1) = \vec{D'} \cdot e^{bl} \cdot \cos(2\pi l) + \vec{X^*}(t)$
16:                     **end if**
17:                     **end if**
18:                     Evaluate new solution
19:                 **end for**
20:                 Compute performance of WOE ($\Delta fit$)
21:                 **if** $\Delta fit < \Omega$ **then** Stop running WOA and switch to AHA  ▷ Detect stagnation issue
22:                 **end if**
23:             **until** $Iter_t < I_{WOA}$
24:             **repeat**  ▷ Run second component (AHA)
25:                 **for** each $Particle$ in $N$ **do**
26:                     Update best-found solution and compute three movement strategies Eq. 14
27:                     **if** $r < 0.5$ **then** $X_i(t+1) = \begin{cases} X_i(t) & \text{if } f(X_i(t)) \leq f(V_i(t+1)) \\ V_i(t+1) & \text{otherwise} \end{cases}$
28:                     **else** $V_i(t+1) = X_{i,t}(t) + b \times D \times X_i(t), b \in N(0,1)$
29:                     **end if**
30:                     Update new particle's position Eq. 19 and Evaluate particle position
31:                     Update particle control parameters
32:                 **end for**
33:                 Compute performance of AHA ($\Delta fit$)
34:                 **if** $\Delta fit < \Omega$ **then** Stop running AHA and switch to CMA  ▷ Detect stagnation issue
35:                 **end if**
36:             **until** $Iter_t < I_{AHA}$
37:             **repeat**  ▷ Run CMA
38:                 Update distribution mean $m_{k+1} = m_k + \sum_{i=1}^{\mu} \beta_i (X_{i:N_\lambda} - m_k)$
39:                 Update $\sigma_k$, cumulative adaptation strategy (CSA) using Eq. 22
40:                 Update evolution path $p_\sigma$ using Eq. 23
41:                 Update covariance matrix
42:                 Update the population $X_i \sim m_k + \sigma_k \times \mathcal{N}(0, C_k)$
43:                 Evaluate population $f(X_i)$
44:             **until** $\Delta fit < \Omega$
45:             Update cumulative-fitness
46:         **end for**
47:     **end for**
48: **until** $Meta_{iter} \leq Max_{iter}$

# 4. Numerical results and discussions

## 4.1. Landscape analysis

The landscape analysis of both Power take-off (PTO) and design parameters plays a vital role in the development and optimisation of total power output in a hybrid platform. This is primarily due to the direct influence of PTO parameters, such as damping, stiffness, and control strategies, on power extraction efficiency from ocean waves [24]. By conducting a comprehensive landscape analysis of these parameters, industries can identify optimal configurations that result in improved power conversion efficiency. This analysis enables the selection of the most suitable PTO parameters tailored to the specific design of a WEC. Additionally, evaluating the performance of different configurations allows for identifying regions in the parameter space that yield the best power-to-weight ratio, optimal power absorption, and minimal structural loads. This knowledge aids in optimising the overall performance of the hybrid platform and guides the design process towards achieving higher power output and efficiency.

In this study, a real-time tuning approach is assumed for both Power take-off (PTO) stiffness and damping, depending on the desired objective, while maintaining identical settings for all three Wave Energy Converters (WECs). However, it is recognized that the values of these parameters may vary for each environmental condition. The objective is to examine how the WEC PTO configuration, with a fixed radius of 5 m and a distance of 20 m to the platform, impacts the key performance criteria of the hybrid wind-wave system. To achieve this, a grid search technique is employed. In this experimental setup, the 20 damping values and stiffness parameters are assumed to be fixed, with the PTO stiffness and damping constrained to positive values only. This investigation aims to provide insights into the impact of different PTO configurations on the performance of the hybrid system.

The landscape analysis results for three fitness functions for Sydney offshore sea site can be seen in Figure 6. The landscape of PTO parameters for Sydney (Figure 6a) sea site demonstrates that there is the same pattern of PTO parameters interactions with power loss coefficient ($1/q$, $log(Power_{isolated}/Power_{hybrid})$). From these landscapes, the regions with the minimum power loss can be subjected between $B_{pto} = [1.0E + 05, 1.0E + 07]$ and $K_{pto} = [1.0E + 01, 1.0E+07]$. These PTO configurations can perform best associated with maximum average annual hybrid power, as can be seen in (Figure 6c). The landscapes achieved based on a coefficient of hybrid nacelle acceleration and FOWT nacelle acceleration show that the proper regions of PTO parameters can be different depending on the wave scenario (See Figure 6b).

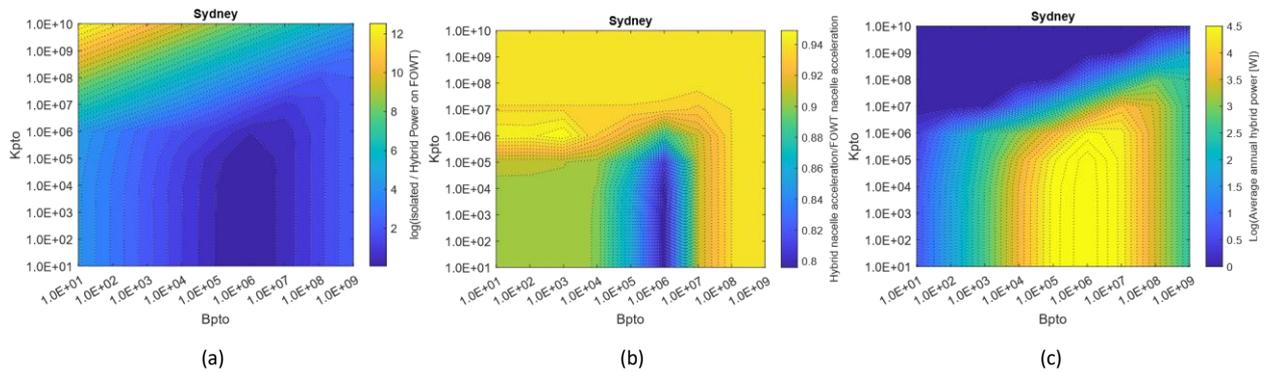

Figure 6: Landscape analysis of PTO parameters ($K_{pto}$ and $B_{pto}$) on the performance of the hybrid wind-wave system based on Sydney offshore site. Two control parameters, distance and WEC radius, were kept fixed at 20 m and 5 m: a and d) isolated WEC power / Power WEC on FOWT (Objective function 1 ($1/q$)), b and e) Hybrid nacelle acceleration / FOWT nacelle acceleration (Objective function 2), and c and f) Average annual hybrid power [W],

## 4.2. Meta-heuristic performances and optimisation findings

In this section, a comprehensive assessment is conducted to evaluate the effectiveness of the proposed ensemble optimiser in maximising the total power output while minimising the nacelle acceleration of the wind turbine. To establish a fair comparison, a diverse set of nine state-of-the-art and widely used swarm and evolutionary-based algorithms is employed. The control parameters used in each optimisation method are derived from authoritative literature sources and are presented in detail in Table 3. The population size and maximum number of evaluations are consistently set at 25 and 1000, respectively, across all optimisation methods. The performance evaluation is based on two critical criteria: the efficiency in swiftly identifying optimal solutions and the reliability of each method in consistently attaining the best-found solution in every run. By considering these metrics, a thorough comparative analysis is conducted, facilitating the assessment of the proposed ensemble optimiser against established other algorithms.

Table 3: The settings of optimisation methods in the context of the hybrid wind-wave converter problem. In CMA-ES, $\mu$, $\lambda$, and $w$ are the numbers of offspring, parents, and the weight of parents, respectively.

| | Abbreviation | Full name | Initial settings |
|---|---|---|---|
| 1 | DE [85] | Differential Evolution | $\beta_{min}$ = 0.2, $\beta_{max}$ = 0.8, $p_{CR}$ = 0.2, $\lambda$ = $Best_{Sol}$ + $\beta \times$ ($Position_{r1}$ − $Position_{r2}$) |
| 2 | CMA-ES [86] | Evolution Strategy with Covariance Matrix Adaptation | $\mu = \lambda/2$, $w = log(\mu + 0.5) - log(1:\mu)$, $w = w/P(w)$, $\mu_{effective} = 1/P(w_2)$, $\sigma_0 = 0.1 \times (UB - LB)$ |
| 3 | PSO [87] | Particle Swarm Optimisation | $\kappa = 1$, $\varphi_1 = 2.05$, $\varphi_2 = 2.05$, $\varphi = \varphi_1 + \varphi_2$, $\chi = 2 \times \kappa/(2 - \varphi - \sqrt{(\varphi^2 - 4*\varphi)})$, $w = \chi$, $w_{damp} = 1$, $c_1 = \chi * \varphi_1$, $c_2 = \chi * \varphi_2$ |
| 4 | GWO [88] | Grey Wolf Optimiser | $\alpha = 1.5 - Iter \times ((1.5)/Max_{Iter})$ |
| 5 | WOA [89] | Whale Optimisation Algorithm | $\beta = 1$, $\alpha = 2 \times exp(-(3 * Iter/Max_{Iter})^2)$, $\alpha_2 = -1 + Iter \times ((-1)/Max_{Iter})$, $L = (\alpha_2 - 1)*rand + 1$ |
| 6 | MPA [90] | Marine Predators Algorithm | $CF = (1 - Iter/Max_{Iter})^{(2 \times Iter/Max_{Iter})}$, $RL = 0.05 \times levy(N_{pop}, dim, 1.5)$, $RB = randn(N_{pop}, dim)$ |
| 7 | AHA [77] | Artificial Hummingbird Algorithm | Predefined parameters |
| 8 | EO [91] | Equilibrium Optimiser | $V = 1$, $\alpha_1 = 2$, $\alpha_2 = 1$, $GP = 0.5$ |
| 9 | SCA [92] | Sine Cosine Algorithm | $\alpha = 2, r = \alpha - Iter * (\alpha/Max_{Iter})$ |

Figure 7 presents the statistical optimisation outcomes achieved by the proposed ensemble algorithm (EEA) in comparison with nine alternative optimisation methods. The evaluation criterion employed is minimising the power loss coefficient (1/q) for wave scenarios in Sydney and Port Lincoln. Each box within the figure represents the interquartile range, encapsulating the central 50% of the optimisation results, with the lower and upper quartiles demarcated by lines. The median is indicated by a clear line. The box length signifies the extent of variability or spreads in the optimisation performance of the algorithms within this range.

Analysis of Figures 7a and 7d reveals the superior performance of EEA in attaining hybrid platform parameters that yield minimum power loss (1/q), irrespective of the specific sea site considered. Notably, the power loss variance exhibited by EEA's best-found solutions is considerably narrower than that of the alternative methods, indicating heightened robustness and reliability in exploring optimal solutions within the complex and multi-modal search space. Furthermore, EEA outperforms the other nine optimisation algorithms in terms of average annual hybrid power metric, showcasing significant improvements across both case studies.

This achievement can be attributed to the combination of WOA and AHA within the hybrid EEA. WOA contributes its inherent efficiency in exploration capabilities, while AHA enhances the adaptability and robustness of the algorithm when confronted with complex search spaces. This amalgamation allows for a comprehensive search space exploration, thereby augmenting the coverage of diverse solutions and increasing the likelihood of discovering optimal outcomes. Moreover, the performance of optimisation methods can be enhanced by transferring a continuous search space to a discretised model. This transformation reduces the search space dimensionality and facilitates a grid-based exploration approach.

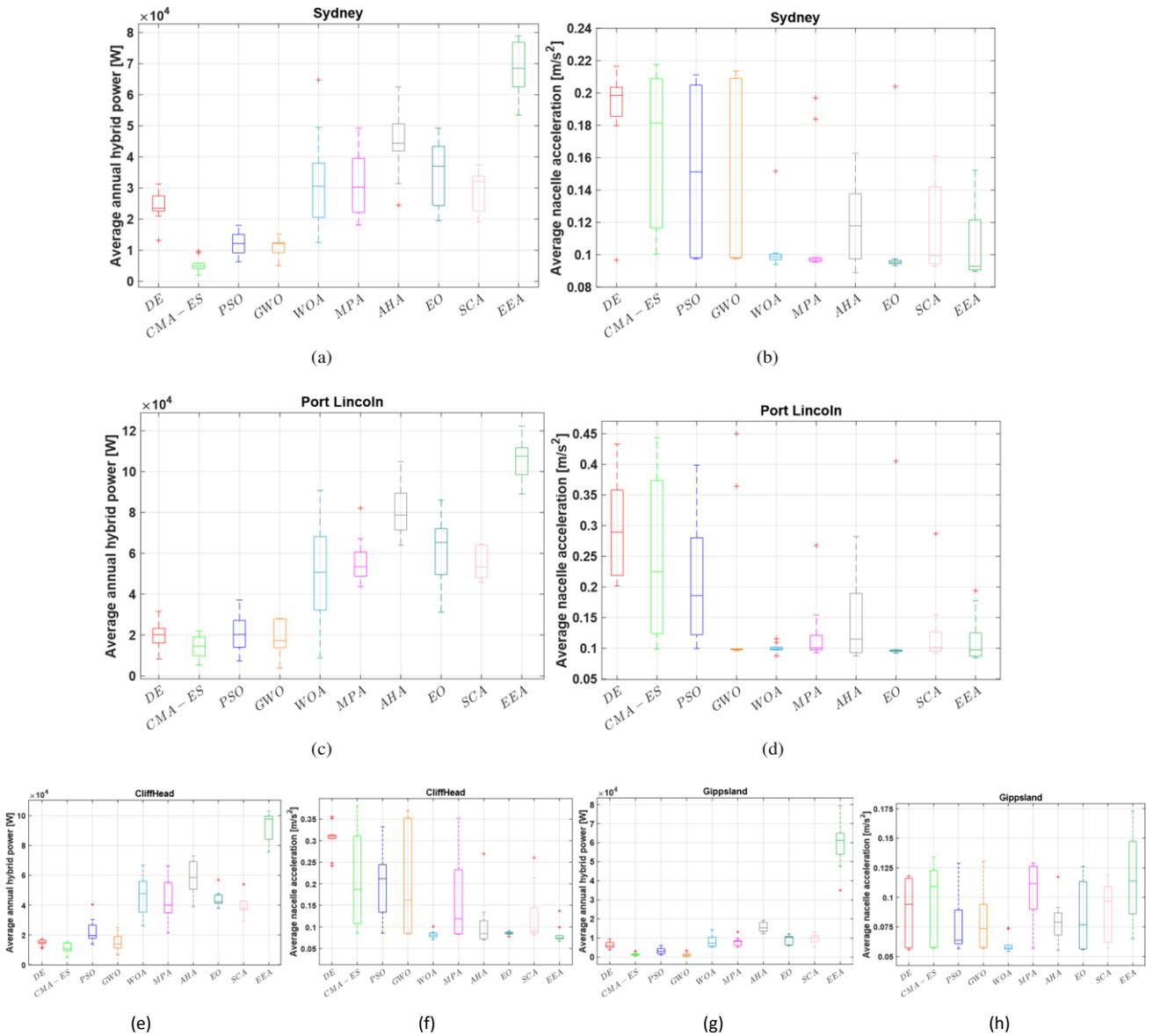

Figure 7: Statistical optimisation results of the hybrid wind-wave system performance under the maximising the Average annual hybrid power [W] and minimising Average nacelle acceleration [$m/s^2$] using nine optimisation methods based on the four wave site. Each optimisation method runs ten times with random initialisation for the first population. The radius of the WECs is fixed at 5 m.

Table A.1 presents additional technical comparison findings between the ensemble algorithm (EEA) and nine alternative evolutionary models. The results highlight the exceptional performance of EEA in statistically minimising the power loss of the hybrid platform. Notably, EEA achieved the highest optimisation performance among all the methods considered. Furthermore, among the nine optimisation algorithms, it is evident that both the AHA and EO exhibited superior performance compared to the remaining techniques. These two methods secured second and third place, respectively, after EEA in both sea sites, further emphasising their effectiveness in optimising power loss.

Table 4: The statistical results conducted on nine optimisation methods to maximise the average annual hybrid power for four sea sites. Each method was independently executed ten times, and the solution with the highest performance was selected from each run. The radius of the WECs is fixed at 5 m.

| | Average annual hybrid power [W], Sydney | | | | | | | | | |
|---|---|---|---|---|---|---|---|---|---|---|
| Metric | DE | CMAES | PSO | GWO | WOA | MPA | AHA | EO | SCA | EEA |
| Min | 1.32E+04 | 1.96E+03 | 6.26E+03 | 5.04E+03 | 1.25E+04 | 1.81E+04 | 2.45E+04 | 1.95E+04 | 1.91E+04 | 5.34E+04 |
| Max | 3.12E+04 | 9.64E+03 | 1.80E+04 | 1.51E+04 | 6.47E+04 | 4.92E+04 | 6.25E+04 | 4.93E+04 | 3.74E+04 | 7.89E+04 |
| Mean | 2.43E+04 | 5.22E+03 | 1.20E+04 | 1.09E+04 | 3.25E+04 | 3.12E+04 | 4.50E+04 | 3.52E+04 | 2.98E+04 | 6.86E+04 |
| Median | 2.35E+04 | 4.77E+03 | 1.21E+04 | 1.22E+04 | 3.06E+04 | 3.02E+04 | 4.44E+04 | 3.70E+04 | 3.21E+04 | 6.84E+04 |
| STD | 5.31E+03 | 2.50E+03 | 3.93E+03 | 3.36E+03 | 1.58E+04 | 1.05E+04 | 1.11E+04 | 1.05E+04 | 6.33E+03 | 8.29E+03 |
| | Average annual hybrid power [W], Port Lincoln | | | | | | | | | |
| Metric | DE | CMAES | PSO | GWO | WOA | MPA | AHA | EO | SCA | EEA |
| Min | 8.27E+03 | 5.30E+03 | 7.29E+03 | 3.78E+03 | 8.79E+03 | 4.37E+04 | 6.40E+04 | 3.12E+04 | 4.58E+04 | 8.90E+04 |
| Max | 3.15E+04 | 2.20E+04 | 3.72E+04 | 2.80E+04 | 9.08E+04 | 8.22E+04 | 1.05E+05 | 8.60E+04 | 6.45E+04 | 1.22E+05 |
| Mean | 2.02E+04 | 1.41E+04 | 2.05E+04 | 1.89E+04 | 5.10E+04 | 5.63E+04 | 8.10E+04 | 6.14E+04 | 5.51E+04 | 1.07E+05 |
| Median | 2.01E+04 | 1.45E+04 | 2.02E+04 | 1.72E+04 | 5.06E+04 | 5.35E+04 | 7.87E+04 | 6.53E+04 | 5.33E+04 | 1.08E+05 |
| STD | 7.47E+03 | 5.84E+03 | 8.78E+03 | 8.60E+03 | 2.54E+04 | 1.14E+04 | 1.38E+04 | 1.77E+04 | 8.16E+03 | 9.96E+03 |
| | Average annual hybrid power [W], CliffHead | | | | | | | | | |
| Metric | DE | CMAES | PSO | GWO | WOA | MPA | AHA | EO | SCA | EEA |
| Min | 1.10E+04 | 4.82E+03 | 1.36E+04 | 6.52E+03 | 2.60E+04 | 2.13E+04 | 3.89E+04 | 3.78E+04 | 2.91E+04 | 7.57E+04 |
| Max | 1.72E+04 | 1.52E+04 | 4.04E+04 | 2.49E+04 | 6.67E+04 | 6.64E+04 | 7.29E+04 | 5.69E+04 | 5.39E+04 | 1.03E+05 |
| Mean | 1.47E+04 | 1.10E+04 | 2.26E+04 | 1.51E+04 | 4.64E+04 | 4.30E+04 | 5.85E+04 | 4.41E+04 | 3.92E+04 | 9.31E+04 |
| Median | 1.49E+04 | 1.05E+04 | 1.94E+04 | 1.37E+04 | 4.76E+04 | 3.98E+04 | 5.84E+04 | 4.21E+04 | 3.80E+04 | 9.78E+04 |
| STD | 2.03E+03 | 3.69E+03 | 7.96E+03 | 5.71E+03 | 1.42E+04 | 1.45E+04 | 1.16E+04 | 5.53E+03 | 6.79E+03 | 9.58E+03 |
| | Average annual hybrid power [W], Gippsland | | | | | | | | | |
| Metric | DE | CMAES | PSO | GWO | WOA | MPA | AHA | EO | SCA | EEA |
| Min | 3.69E+03 | 4.35E+02 | 1.09E+03 | 1.05E+02 | 5.20E+03 | 5.14E+03 | 1.21E+04 | 5.92E+03 | 5.09E+03 | 3.50E+04 |
| Max | 9.38E+03 | 3.03E+03 | 6.12E+03 | 3.41E+03 | 1.43E+04 | 1.32E+04 | 1.92E+04 | 1.21E+04 | 1.32E+04 | 7.95E+04 |
| Mean | 6.19E+03 | 1.36E+03 | 3.16E+03 | 1.19E+03 | 8.20E+03 | 8.01E+03 | 1.55E+04 | 9.29E+03 | 9.70E+03 | 5.92E+04 |
| Median | 5.88E+03 | 1.28E+03 | 3.06E+03 | 1.02E+03 | 7.34E+03 | 8.17E+03 | 1.54E+04 | 1.03E+04 | 1.01E+04 | 6.12E+04 |
| STD | 1.67E+03 | 7.28E+02 | 1.61E+03 | 1.02E+03 | 3.02E+03 | 2.40E+03 | 2.48E+03 | 2.29E+03 | 2.41E+03 | 1.19E+04 |

Based on the observations derived from Figure 8, the ensemble algorithm (EEA) emerges as the top-performing optimiser, exhibiting rapid convergence towards the optimal configuration and maintaining a consistently steep slope throughout the optimisation process in both sea sites. Moreover, EEA demonstrates high convergence stability, primarily attributed to its robust exploration behaviour in the initial generations. It is followed by a smart exploitation strategy that effectively prevents convergence to local optima, thanks to its novel search characteristics. In contrast, the WOA initially displayed a notable convergence speed among the remaining eight optimisers, surpassing the others. However, it encountered the issue of premature convergence, hindering the improvement of solution quality in subsequent iterations. In both wave scenarios depicted in Figure 8a and b, the second-best convergence speed, resulting in the continuous enhancement of the hybrid model's performance in terms of average hybrid power, was exhibited by the Artificial Hummingbird Algorithm (AHA).

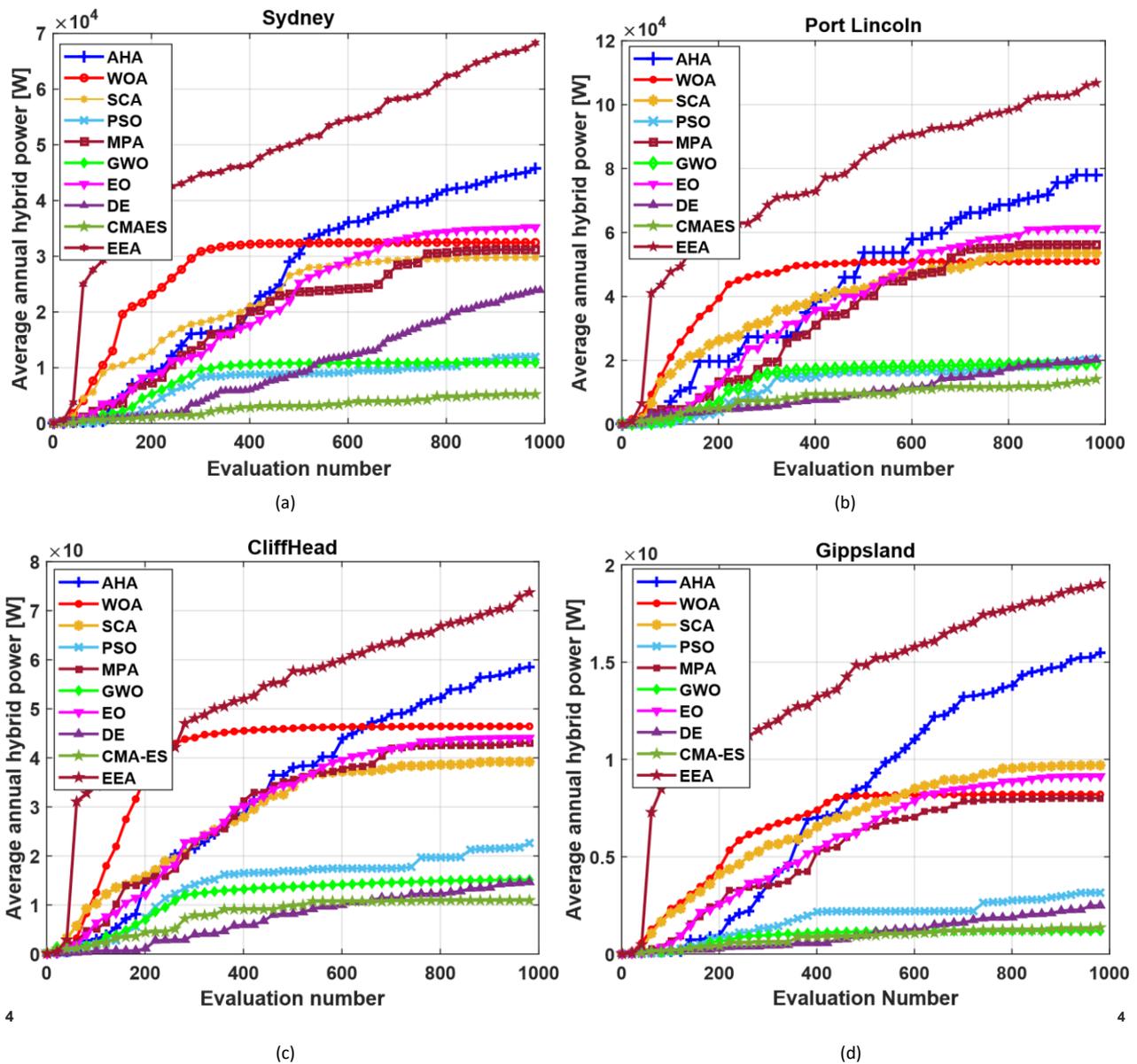

Figure 8: A comparative analysis to assess the convergence speed of nine meta-heuristics for a) Sydney and b) Port Lincoln sea site in the context of maximising the hybrid power output of a WEC while simultaneously minimising nacelle acceleration. The radius of the WECs is fixed at 5 m.

Figure A.3 presents a comparative analysis of the power loss convergence speed between the proposed ensemble optimiser and nine other optimisation methods in the Sydney (a) and Port Lincoln (b) sites. The results reveal that optimising the hybrid model's performance based on the Sydney wave scenarios poses more significant challenges compared to Port Lincoln, leading to subpar performance for some popular optimisers such as Differential Evolution (DE), Covariance Matrix Adaptation Evolution Strategy (CMA-ES), and Particle Swarm Optimisation (PSO). In contrast, the ensemble algorithm (EEA) exhibits rapid convergence behaviour towards subregions with high-quality solutions, utilising only 20% of the computation runtime. Subsequently, EEA maintains a smooth convergence curve to consistently enhance the fitness of the discovered solutions within the remaining budget. Like the power hybrid convergence plot in Figure 8, the Whale Optimization Algorithm (WOA) and Sine Cosine Algorithm (SCA) demonstrate competitive convergence rates during the initial iterations, outperforming other methods, except for EEA. However,

the presence of local optima hampers their ability to explore new areas of the search space, necessitating adjustments or enhancements to overcome stagnation.

Tables 5 and 6 report the optimal Power Take-Off (PTO) parameters and corresponding platform configurations obtained by the ensemble algorithm (EEA) and nine widely used optimisation methods. These results are based on minimising the power loss coefficient in the Sydney and Port Lincoln Sea sites, respectively. A noteworthy observation is that in the Sydney case study, the optimisation framework, including EEA, was unable to identify a solution with a power loss coefficient of less than 1. This finding indicates the challenging nature of the optimisation problem in this particular location. However, Table 5 demonstrates that EEA successfully proposed a specific PTO configuration and distance that achieved a low power loss coefficient of 0.99 in the Port Lincoln case study.

Furthermore, the best-found solution by EEA suggests a distance variable of 20 m and 40 m in the Sydney and Port Lincoln case studies, respectively. Lastly, it is evident that the power loss coefficient values obtained for the Port Lincoln site are consistently lower than those for the Sydney site on average. This discrepancy highlights the varying levels of power loss between the two locations and underscores the importance of site-specific optimisation approaches. To make a long story short, Tables 5 and 6 provide detailed insights into the optimal PTO parameters and platform configurations obtained by EEA and other optimisation methods in the Sydney and Port Lincoln Sea sites. While the optimisation framework struggled to achieve power loss coefficients below 1 in Sydney, EEA successfully proposed efficient configurations with improved performance in the Port Lincoln case study. Additionally, the distance variable in the best-found solutions varied between the two sites, emphasising the optimisation process's site-specific nature.

Table 5: Best-found configurations unearthed by the optimisation methods for the Sydney site. The radius of the WECs is fixed at 5 m.

| Parameter | DE | CMAES | PSO | GWO | WOA | MPA | AHA | EO | SCA | EEA |
|---|---|---|---|---|---|---|---|---|---|---|
| Average annual hybrid power [W] | 3.12E+04 | 9.64E+03 | 1.80E+04 | 1.51E+04 | 6.47E+04 | 4.92E+04 | 6.25E+04 | 4.93E+04 | 3.74E+04 | 7.89E+04 |
| Distance | 5.00E+01 | 5.00E+01 | 5.00E+01 | 2.00E+01 | 5.00E+01 | 5.00E+01 | 2.00E+01 | 2.00E+01 | 5.00E+01 | 2.00E+01 |
| $\sum_{i=1}^{N_k} \text{Kpto}_i / N_k$ | 3.85E+09 | 3.56E+09 | 3.84E+09 | 2.88E+09 | 7.86E+07 | 1.01E+09 | 5.83E+06 | 1.99E+09 | 7.56E+08 | 2.92E+06 |
| $\sum_{i=1}^{N_B} \text{Bpto}_i / N_B$ | 3.23E+09 | 5.19E+09 | 4.41E+09 | 3.06E+09 | 6.70E+06 | 1.53E+09 | 6.42E+08 | 1.19E+09 | 5.22E+08 | 1.49E+08 |

Table 6: Best-found configurations unearthed by the optimisation methods for the Port Lincoln site. The radius of the WECs is fixed at 5 m.

| Parameter | DE | CMAES | PSO | GWO | WOA | MPA | AHA | EO | SCA | EEA |
|---|---|---|---|---|---|---|---|---|---|---|
| Average annual hybrid power [W] | 3.15E+04 | 2.20E+04 | 3.72E+04 | 2.80E+04 | 9.08E+04 | 8.22E+04 | 1.05E+05 | 8.60E+04 | 6.45E+04 | 1.22E+05 |
| Distance | 3.50E+01 | 3.00E+01 | 5.00E+01 | 2.00E+01 | 2.00E+01 | 2.00E+01 | 4.00E+01 | 2.00E+01 | 4.00E+01 | 4.00E+01 |
| $\sum_{i=1}^{N_k} \text{Kpto}_i / N_k$ | 3.84E+09 | 3.58E+09 | 4.01E+09 | 3.51E+09 | 3.73E+06 | 3.52E+08 | 5.86E+05 | 8.21E+08 | 1.04E+09 | 3.45E+07 |
| $\sum_{i=1}^{N_B} \text{Bpto}_i / N_B$ | 4.63E+09 | 4.20E+09 | 3.13E+09 | 3.63E+09 | 6.26E+07 | 9.04E+08 | 2.81E+08 | 1.43E+09 | 6.04E+08 | 4.28E+07 |

These observations shed light on the comparative performance of the optimisers, with EEA showcasing exceptional convergence speed, stability, and exploration-exploitation balance. AHA also demonstrates commendable convergence speed, contributing to the steady improvement of the hybrid model's performance.

*4.3. Developed hybrid model findings*

In this section, we present the optimisation results of the extended hybrid wind-wave model, incorporating the radius of the wave energy converter (WEC) as an additional optimisation parameter. To evaluate the performance of the proposed algorithm (EEA), a comparative framework was developed, as illustrated in Figure 9. Consistent with previous findings, EEA demonstrates exceptional performance in maximising the absorbed power of the hybrid system. It outperforms other algorithms, achieving improvements over WOA, AHA, DE, PSO, EFADE, LSHADE-cnEpSin, SaDPSO, SaNSDE, and SLPSO by 236.85%, 110.58%, 438.54%, 3124.81%, 1220.31%, 497.88%, 2182.43%, 638.43%, and 349.47%, respectively. These results highlight the superior capability of EEA in optimising complex hybrid energy systems.

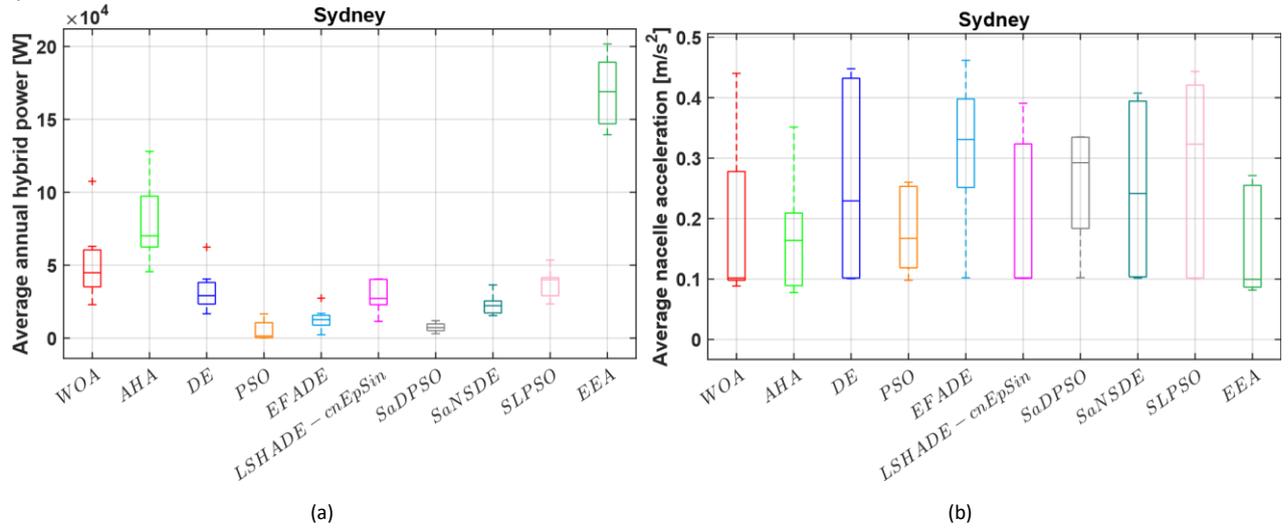

Figure 9: Statistical optimisation results of the hybrid wind-wave system performance under the maximising the average annual hybrid power using nine optimisation methods based on the Sydney with considering the radius of wave converters as an optimisation parameter. Each optimisation method runs ten times with random initialisation for the first population.

Table 7 presents the detailed statistical optimization results of EEA compared to four popular meta-heuristics and five advanced optimisation methods. EEA demonstrated superior average performance, validating its effectiveness and robustness across ten runs with random initialisation of the initial population.

Table 7: The statistical results were conducted on nine optimisation methods to maximise the average annual hybrid power for Sydney sea sites with radius parameter. Each method was independently executed ten times, and the solution with the highest performance was selected from each run.

| Average annual hybrid power [W], Sydney | | | | | | | | | | |
|---|---|---|---|---|---|---|---|---|---|---|
| Metric | WOA | AHA | DE | PSO | EFADE | LSHADE-cnEpSin | SaDPSO | SaNSDE | SLPSO | EEA |
| Min | 2.30E+04 | 4.56E+04 | 1.67E+04 | 1.55E+02 | 2.33E+03 | 1.15E+04 | 3.07E+03 | 1.54E+04 | 2.35E+04 | 1.39E+05 |
| Max | 1.07E+05 | 1.28E+05 | 6.24E+04 | 1.66E+04 | 2.74E+04 | 4.05E+04 | 1.20E+04 | 3.65E+04 | 5.35E+04 | 2.02E+05 |
| Mean | 5.02E+04 | 8.03E+04 | 3.14E+04 | 5.24E+03 | 1.28E+04 | 2.83E+04 | 7.40E+03 | 2.29E+04 | 3.76E+04 | 1.69E+05 |
| Median | 4.48E+04 | 7.01E+04 | 2.91E+04 | 1.46E+03 | 1.27E+04 | 2.72E+04 | 7.22E+03 | 2.23E+04 | 4.01E+04 | 1.69E+05 |
| STD | 2.38E+04 | 2.67E+04 | 1.33E+04 | 6.05E+03 | 6.90E+03 | 9.92E+03 | 2.78E+03 | 6.67E+03 | 9.61E+03 | 2.11E+04 |
| Average nacelle acceleration [$m/s^2$], Sydney | | | | | | | | | | |
| Metric | WOA | AHA | DE | PSO | EFADE | LSHADE-cnEpSin | SaDPSO | SaNSDE | SLPSO | EEA |
| Min | 8.84E-02 | 7.77E-02 | 1.01E-01 | 9.80E-02 | 1.02E-01 | 1.01E-01 | 1.02E-01 | 1.02E-01 | 1.01E-01 | 8.17E-02 |
| Max | 4.40E-01 | 3.51E-01 | 4.48E-01 | 2.60E-01 | 4.62E-01 | 3.91E-01 | 3.35E-01 | 4.07E-01 | 4.43E-01 | 2.71E-01 |
| Mean | 1.80E-01 | 1.79E-01 | 2.62E-01 | 1.77E-01 | 3.20E-01 | 1.78E-01 | 2.57E-01 | 2.44E-01 | 2.78E-01 | 1.51E-01 |
| Median | 1.02E-01 | 1.64E-01 | 2.29E-01 | 1.67E-01 | 3.31E-01 | 1.02E-01 | 2.92E-01 | 2.42E-01 | 3.23E-01 | 9.96E-02 |

| STD | 1.39E-01 | 9.83E-02 | 1.61E-01 | 6.29E-02 | 1.01E-01 | 1.18E-01 | 8.76E-02 | 1.23E-01 | 1.51E-01 | 8.27E-02 |

Figure 10 is a parallel coordinates plot visualizing multi-dimensional data, where each data point is represented as a line traversing multiple vertical axes: (a) $B_{pto}$, (b) $K_{pto}$, and (c) $a$ and $L$. This visualization aids in identifying relationships, patterns, clusters, and outliers in high-dimensional data. The dark red lines highlight the best-performing hybrid system's parameters, showcasing complex interactions among PTO parameters and the largest radius size, with distances exceeding 40 m.

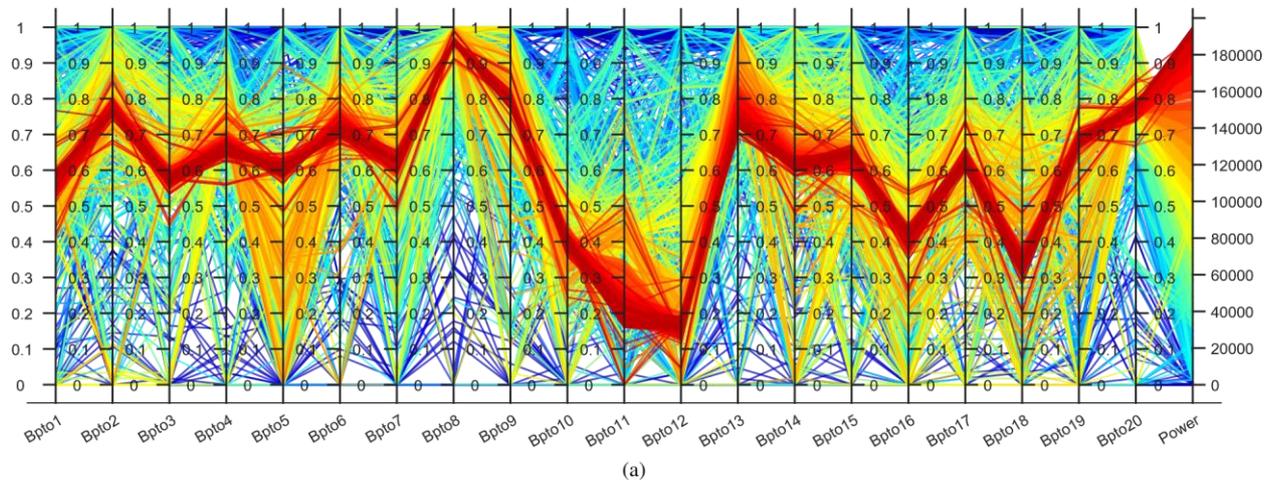

(a)

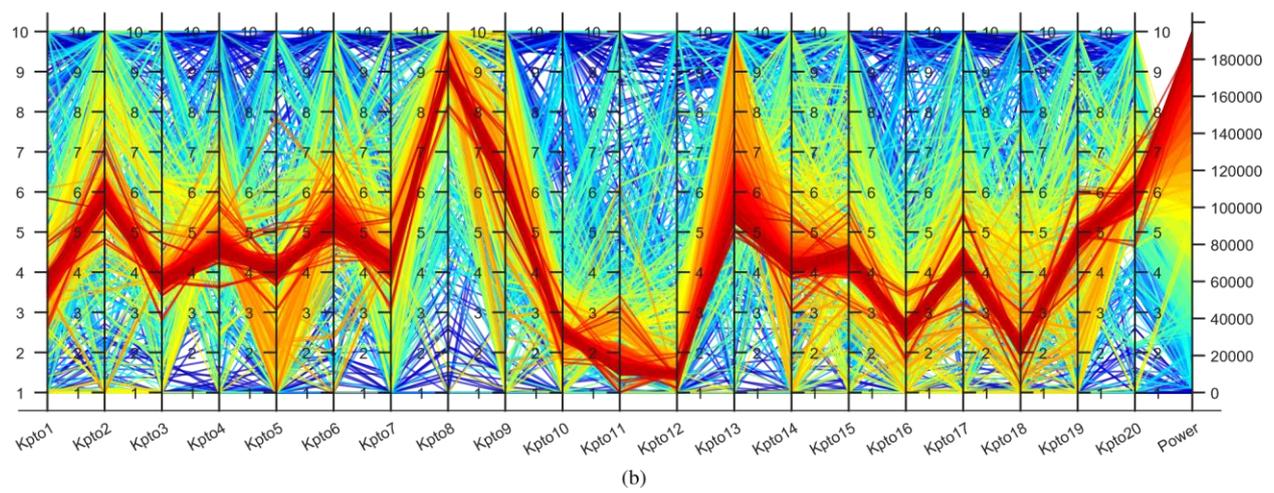

(b)

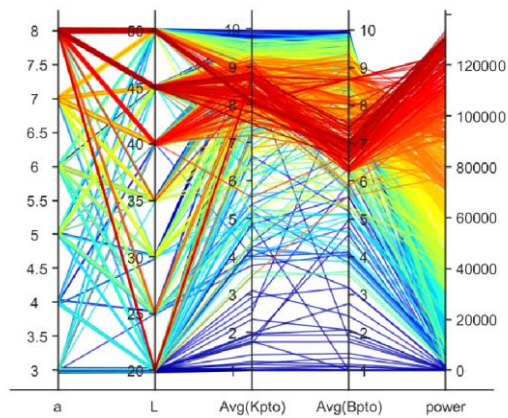

(c)

Figure 10: The optimization process of the proposed ensemble method aimed at maximizing the hybrid power of the system. The plot represents 20 variables corresponding to the $K_{pto}$ parameters and 20 variables for the $B_{pto}$ parameters, alongside two critical design parameters: *a* (radius of the wave converters) and *L* (distance). Each line in the plot corresponds to a different evaluation of the optimization algorithm, highlighting the relationships and interactions between the various parameters throughout the optimization iterations.

## 5. Conclusions

Harnessing renewable energy is pivotal in the fight against climate change and global warming, as highlighted by the Paris Agreement, by curtailing greenhouse gas emissions and shifting towards sustainable energy frameworks to cap global temperature rise [93]. Hybrid wave-wind energy systems diversify sources, enhancing resilience and reducing supply risks. Integration of wave and wind energy yields consistent output, smoothing variability and ensuring reliable power. However, optimising hybrid systems is complex due to intricate physical interactions, resulting in a heterogeneous optimisation space. To address the challenges mentioned earlier, we present an innovative ensemble optimisation technique (EEA), coupled with a Novelty search, Covariance matrix adaptation, and Discretisation approach to preserve solution diversity and expedite the optimisation process in optimising the performance of the hybrid wind-wave system based on four offshore Australian sites. Our approach concurrently considers both geometry and Power Take-Off (PTO) parameters to maximise the average power output of wind and wave systems. This approach maintains solution diversity, mitigates the curse of dimensionality, and accelerates the optimisation process. In comparison to nine established optimisation methods, the proposed ensemble optimiser (EEA) demonstrates remarkable improvements in average absorbed hybrid power. In the Port Lincoln Sea site (radius is fixed at 5 m), EEA outperforms WOA, EO, and AHA by 109%, 74%, and 32%, respectively. Similarly, in the Sydney Sea site, EEA surpasses WOA, EO, and AHA by 111%, 95%, and 52%, respectively. Additionally, EEA exhibits substantial performance in reducing the average power loss coefficient. These comparative findings strongly validate EEA's superiority in terms of configuration, quality and convergence rate.

Future research will focus on enhancing hybrid system design and advancing optimisation methods to address challenges in renewable energy systems. Efforts will include refining wave-wind integration models and developing adaptive algorithms for dynamic environmental conditions. Incorporating deep learning into the optimisation framework will further improve adjustments to geometry and Power Take-Off (PTO) parameters, maximising power output and efficiency. These advancements aim to achieve more reliable and sustainable energy systems aligned with global climate goals.

**Acknowledgment**

The authors thank Dr Marielle de Oliveira for providing drawings of the hybrid wind-wave system. This research is funded by the Australia-China Science and Research fund, the Australian Department of Industry, Innovation and Science (ACSRF66211), and the Ministry of Science and Technology of China (2017YFE0132000). Author Sergiienko is the recipient of an Australian Research Council Early Career Industry Fellowship (project number IE230100545) funded by the Australian Government.